%% file: main.tex
\documentclass[runningheads]{llncs}
\usepackage{graphicx}
\usepackage{listings}
\usepackage{float}
\input{preample}

\input{notations}

\begin{document}
%
\title{Spatiotemporal Cardiac Statistical Shape Modeling: A Data-Driven Approach}
\titlerunning{Spatiotemporal Cardiac SSM}
\author{Jadie Adams\thanks{Contributed equally.}\inst{1,2}  \and
Nawazish Khan$^\star$\inst{1,2} \and Alan Morris\inst{1,2} \and
Shireen Elhabian\inst{1,2} }

\authorrunning{Adams and Khan et al.}
\institute{Scientific Computing and Imaging Institute, University of Utah, UT, USA \and
School of Computing, University of Utah, UT, USA \\
\email{ \{jadie, nawazish.khan, amorris, shireen \}@sci.utah.edu }
}

%
\maketitle

\input{abstract}
\input{intro}
\input{methods}
\input{results}
\input{conclusion}
%
%
\bibliographystyle{splncs04}
\bibliography{references}

\end{document}


\appendix 

\section{Linear Dynamical System}
\subsection{Time-Variant Model}

A linear dynamical system (LDS) is a generative model capable of modeling time-series observations (Fig. 1). LDS is state space model that differs from a Hidden Markov Model (HMM) in two ways: (1) it employs linear Gaussian dynamics and measurements and (2) it uses a continuous latent/state variable instead of discrete.
LDS makes use of a latent representation ($\ldsS$) and is defined by equations \ref{eq:state} - \ref{eq:prior}.

\begin{align}
    \text{State Equation: } \quad \s_{n,t} &= \A_{t}\s_{n,t-1} + \boldepsilon_{n,t}^s & \text{where} \ \boldepsilon_{n,t}^s \in \mathcal{N}(0,\boldSigma^s)
    \label{eq:state} \\
    \text{Observation Equation: } \quad \x_{n,t} &= \W_{t}\s_{n,t} + \boldepsilon_{n,t}^x & \text{where} \ \boldepsilon_{n,t}^x \in \mathcal{N}(0,\boldSigma^x) 
    \label{eq:obs} \\
    \text{Prior Equation: } \quad \s_{n,1} &= \boldmu_{0} + \boldphi_{n,0} & \text{where} \ \boldphi_{n,0} \in \mathcal{N}(0,\V_{0})
    \label{eq:prior}
\end{align}
Here $\s_{n,t}\in \mathbb{R}^{L}$ is the state (latent) vector, $\A_t \in \mathbb{R}^{L \times L}$ is the transition matrix, $\boldSigma^s \in \mathbb{R}^{L \times L}$ is the state covariance matrix, $\W_t\in \mathbb{R}^{dM \times L}$ is the loading/observation-system matrix, $\boldSigma^x \in \mathbb{R}^{dM \times dM}$ is the observation covariance matrix,  $\boldmu_0 \in \mathbb{R}^{L}$ is the latent prior mean, and $\V_0  \in \mathbb{R}^{L \times L}$ is the latent prior covariance matrix.
Thus the time-variant LDS parameters are: $\btheta = \{\{\A_{t}, \W_{t} \}^T_{t=1}, \boldSigma^s, \boldSigma^x, \boldmu_{0},\V_{0} \}$. 

In our case, $\s_{n,t}$ is the $L$ dimensional latent representation of the particle set for the $n^{th}$ subject at time $t$ ($\s_{n,t} \in \mathbb{R}^{L}$). 

\subsection{EM Algorithm}
The latent states and parameters of LDS can be fit using the EM algorithm. The formulation for time-variant LDS with multiple observations is outlined below.

\subsubsection{E-step} \label{E-step}
In the E-step we run the inference algorithm to determine
the posterior distribution of the latent variables $p(\ldsS|\X, \btheta^{old})$. This step is composed of two parts: Kalman filtering and RTS smoothing. 
\begin{enumerate}
    \item \underline{Kalman Filtering}

    Prediction Step: We get the predicted distribution (Eq. \ref{eq:pred_dist}) by solving equations \ref{eq:pred_mu} and \ref{eq:pred_var}.
    \begin{equation}
        P (\s_{n,t} |\x_{n,1:t-1} )  \mathcal{N}(\s_{n,t} | \boldmu^{1:t-1}_{n,t}, \V^{1:t-1}_{t})
        \label{eq:pred_dist}
    \end{equation}
    \begin{equation}
        \boldmu^{1:t-1}_{n,t} = \A_t\boldmu^{1:t-1}_{n,t-1}
        \label{eq:pred_mu}
    \end{equation}
    \begin{equation}
        \V^{1:t-1}_t = \A_t\V^{1:t-1}_{t-1}(\A_t)^{\top} + \boldSigma^s
        \label{eq:pred_var}
    \end{equation}
    
    Measurement Step: We get the filtered distribution (Eq. \ref{eq:filter_dist}) by solving \ref{eq:filter_mu} and \ref{eq:filter_var}.
    \begin{equation}
        p(\s_{n,t} |\x_{n,1:t} ) = \mathcal{N}(\s_{n,t} | \boldmu^{1:t}_{n,t}, \V^{1:t}_{t})
        \label{eq:filter_dist}
    \end{equation}
    \begin{equation}
        \boldmu^{1:t}_{n,t} = \boldmu^{1:t-1}_{n,t} + \K_t(\mathbf{r}_{n,t})
        \label{eq:filter_mu}
    \end{equation}
    \begin{equation}
        \V^{1:t}_t = \V^{1:t-1}_{t} - \K_t\W_t\V^{1:t-1}_t
        \label{eq:filter_var}
    \end{equation}
    Where $ \mathbf{r}_{n,t} = \x_{n,t} - \hat{\x}_{n,t}$ and $\hat{\x}_{n,t} = \W_t\boldmu^{1:t-1}_{n,t}$ and the Kalman Gain matrix is defined as 
    $\K_{t} = \V^{1:t-1}_{t}(\W_t)^\top(\boldP_{t})^{-1}$ 
    where 
    $\boldP_{t} = \W_t\V^{1:t-1}_{t}(\W_t)^\top + \boldSigma^x$.
    
    To remove the risk of computational round-off error causing negative diagonal values, we use an equivalent form of Equation \ref{eq:filter_var}, called the Joseph form:
        \begin{equation}
        \V^{1:t}_t = [\I - \K_t\W_t]\V^{1:t-1}_{t}[\I - \K_t\W_t]^\top + \K_t\boldSigma^x\K_t^\top
        \label{eq:jospeh_filter_var}
    \end{equation}
    
    \item \underline{RTS Smoothing}
    
    We get the smoothed posterior distribution (Eq. \ref{eq:smooth_dist}) by solving equations \ref{eq:smooth_mu} and \ref{eq:smooth_var}.
    \begin{equation}
        p(\s_{n,t} |\x_{n,1:T} ) = \mathcal{N}(\s_{n,t} | \boldmu^{1:T}_{n,t}, \V^{1:T}_{t})
        \label{eq:smooth_dist}
    \end{equation}
    \begin{equation}
        \boldmu^{1:T}_{n,t} = \boldmu^{1:t}_{n,t} + \J_{t}(\boldmu^{1:T}_{n,t+1} - \boldmu^{1:t}_{n,t+1})
        \label{eq:smooth_mu}
    \end{equation}
    \begin{equation}
        \V^{1:T}_{t} = \V^{1:t}_{t} + \J_{t}(\V^{1:T}_{t+1} - \V^{1:t}_{t+1})(\J_{t})^{\top}
        \label{eq:smooth_var}
    \end{equation}
    Where the backward Kalman gain matrix is defined as:
    \begin{equation}
        \J_{t} = \V^{1:t}_{t}(\A_{t+1})^{\top}(\V^{1:t}_{t+1})^{-1}.
    \end{equation}
\end{enumerate}

\subsubsection{M-step}
Now that we can estimate $p(\ldsS|\X, \btheta^{old})$, we optimize $\btheta^{new}$ by maximizing $\mathcal{Q}$:
\begin{align}
    Q( \btheta^{new}, \btheta) = \mathbb{E}_{\mathcal{Z} | \mathcal{X}, \theta}[\ln p(\mathcal{X}, \mathcal{Z} | \btheta )] = \frac{1}{N}\sum_{n=1}^N \mathbb{E}_{\ldsS_n | \X_n, \btheta}[\ln p(\X_n, \ldsS_n | \btheta )]
    \label{eq:EM}
\end{align} 
Where the log likelihood function for a single sample $n$ is:
\begin{align}
    \ln p(\X_{n}, \ldsS_{n} | \theta) = 
    \ln p(\s_{n,1} |\boldmu_{0},\V_{0}) 
    + \sum^T_{t=2}\ln p(\s_{n,t} | \s_{n,t-1}, \A_t,\boldSigma^s) \\ \nonumber
    + \sum^T_{t=1}\ln p(\x_{n,t} | \s_{n,t}, \W_t,\boldSigma^x)
    \label{eq:LL}
\end{align}
 
 


%% file: preample.tex
\usepackage{graphicx}
\usepackage{tikz}
\usetikzlibrary{positioning}
\definecolor {processblue}{cmyk}{0.96,0,0,0}
\usepackage{subfigure}
\usepackage{xcolor}
\usepackage{cite}
\usepackage{xspace} 
\usepackage{amsmath}
\usepackage{amsfonts}

\usepackage{appendix}
\makeatletter
\def\@seccntformat#1{\@ifundefined{#1@cntformat}%
   {\csname the#1\endcsname\quad}  
   {\csname #1@cntformat\endcsname}
}
\let\oldappendix\appendix 
\renewcommand\appendix{%
    \oldappendix
    \newcommand{\section@cntformat}{\appendixname~\thesection\quad}
}
\makeatother
\usepackage{cleveref} 

\newcommand{\ie}{i.e.,~}

%% file: notations.tex
\newcommand{\x}{\mathbf{x}}

\newcommand{\z}{\mathbf{z}}
\newcommand{\s}{\mathbf{s}}

\newcommand{\X}{\mathbf{X}}
\newcommand{\Y}{\mathbf{Y}}
\newcommand{\Z}{\mathbf{Z}}
\newcommand{\ldsS}{\mathbf{S}}

\newcommand{\T}{\mathbf{T}}

\newcommand{\A}{\mathbf{A}}
\newcommand{\W}{\mathbf{W}}

\newcommand{\I}{\mathbf{I}}

\newcommand{\boldSigma}{\mathbf{\Sigma}}

%% file: abstract.tex
\begin{abstract}
Clinical investigations of anatomy's structural changes over time could greatly benefit from population-level quantification of shape, or spatiotemporal statistic shape modeling (SSM).
Such a tool enables characterizing patient organ cycles or disease progression in relation to a cohort of interest. 
Constructing shape models requires establishing a quantitative shape representation (e.g., corresponding landmarks). 
Particle-based shape modeling (PSM) is a data-driven SSM approach that captures population-level shape variations by optimizing landmark placement. However, it assumes cross-sectional study designs and hence has limited statistical power in representing shape changes over time. 
Existing methods for modeling spatiotemporal or longitudinal shape changes require predefined shape atlases and pre-built shape models that are typically constructed cross-sectionally.
This paper proposes a data-driven approach inspired by the PSM method to learn population-level spatiotemporal shape changes directly from shape data. 
We introduce a novel SSM optimization scheme that produces landmarks that are in correspondence both across the population (inter-subject) and across time-series (intra-subject). 
We apply the proposed method to 4D cardiac data from atrial-fibrillation patients and demonstrate its efficacy in representing the dynamic change of the left atrium. 
Furthermore, we show that our method outperforms an image-based approach for spatiotemporal SSM  with respect to a generative time-series model, the Linear Dynamical System (LDS). 
LDS fit using a spatiotemporal shape model optimized via our approach provides better generalization and specificity, indicating it accurately captures the underlying time-dependency. 

\keywords{Statistical Shape Modeling  \and Cardiac Dynamics \and Statistical Morphology Analysis}
\end{abstract}

%% file: intro.tex
\section{Introduction}

Clinical studies that track dynamic or evolving organ shape within and across subjects in a cohort are critical to cardiac research. 
Statistical shape modeling (SSM) from medical image data is a valuable tool in such studies, as it allows for population-level quantification and analysis of anatomical shape variation.
However, well-established approaches for SSM \cite{ShapeWorks, spharm-pdm, Deformetrica} assume static shape and are incapable of directly modeling a population of sequences of shape over time (\ie spatiotemporal data), preventing their use in dynamic or longitudinal analysis.
In SSM, shape is either represented explicitly as dense sets of correspondence points \cite{RTW:Tho17, sarkalkan2014statistical, zachow2015computational}, or implicitly via deformation fields (coordinate transformations in relation to a predefined atlas) \cite{miller2014diffeomorphometry, cootes2004diffeomorphic}.
Progress toward spatiotemporal SSM has been made using both representations. 
Concerning the deformation-based approach, early work applied pure regression, relying on either a cross-sectional atlas for correlation analysis \cite{mansi2009statistical} or techniques for estimating time-varying deformations 
\cite{craene2009large, grenander2007pattern, trouve2012shape, fishbaugh2011estimation}.
Such approaches average shape evolution's without considering the inter-subject variability.
%
Works such as \cite{Durrleman2012TowardAC, fishbaugh2012analysis} require predefined spatiotemporal atlas of normative anatomical changes to quantify patient-specific disease-relevant changes.

The explicit landmark or correspondence-based approach represents shapes in the form of point distribution models (PDM). 
In this paper, we focus on PDM because it is more intuitive than implicit representations, more easily interpreted by clinicians, and can readily be visualized.
Automatic PDM construction has been formulated as an optimization problem via metrics such as entropy \cite{cates2007shape} and minimum description length \cite{davies2002minimum}, as well as using a parametric representation of the surface using spherical harmonics, assuming a sphere template \cite{styner2006spharm}.
The particle-based shape modeling (PSM) approach \cite{cates2007shape} is a data-driven SSM approach for generating PDMs that does not require an initial atlas and learns directly from shape data by optimizing the landmarks placement. 
PSM has proven effective in quantifying group differences and in downstream tasks such as pathology detection and disease diagnosis \cite{bhalodia2020quantifying, harris2013cam, atkins2017quantitative, gaffney2019statistical}. 

However, PSM, like other well-established SSM approaches, assumes static shape and is incapable of directly modeling a population of sequences of shape over time (\ie spatiotemporal data). 
Many methods have been proposed for the statistical analysis of cross-sectional time-series data \cite{davis2010population, hart2010dti, khan2008representation}, including a PSM approach, where a time-agnostic PDM is fit then regressed over time \cite{datar2009particle}.
These approaches either do not contain repeated subject measurements or incorrectly assume shape samples are drawn independently and do not consider the inherent correlation of shapes from the same sequence, confounding the population analysis \cite{gerig2016longitudinal, fitzmaurice2008primer}. 
In other existing methods, an auxiliary time-dependent model such as mixed-effects model \cite{datar2012mixed} is applied to a PDM. 
However, mixed-effects models do not explicitly model dynamically changing anatomies with spatiotemporal movement.
An image-based approach has been proposed for estimating organ segmentation and functional measurements over time \cite{morris2020image}. 
Such an approach could be used for spatiotemporal SSM by first generating a PDM for a single time point across subjects, then independently propagating the correspondence points across individual time sequences using image-based deformable registration.
While this technique incorporates time-dependency into the PDM, it relies on image-based features alone to do so and thus is prone to miscorrespondence. 
We use this approach as a baseline for comparison.

We propose a novel entropy-based PSM optimization objective that disentangles subject and time-dependencies to provide more accurate spatiotemporal SSM. 
This technique encourages maximal inter-subject shape correspondence across the population and temporal intra-subject correspondence across time points.
We demonstrate that the PDM resulting from proposed method effectively captures dynamic shape change over time in a population of left atrium. 
This use case exemplifies dynamic motion over a short discretized time interval, however our method could also be applied to longitudinal studies where sparse observations are consistently made over long periods of time.
%
The existing literature lacks analysis metrics for quantifying how well time-dependency is captured.
To address this, we employ a generative time-series model for verification - the linear dynamical system (LDS).
LDS is a state space model that explicitly represents the underlying evolving process of dynamically changing anatomies over time.
LDS enables PCA-like analysis over time using linear transitions and Gaussian latent variables \cite{Shumway1982, Shumway1991, Switching1994}.
When fit to time-series data via the Expectation-Maximization (EM) algorithm, LDS captures the underlying time evolution and measurement processes and can be used to infer future and missing time points.
We demonstrate that LDS fit to a PDM generated via our approach is more accurate in terms of generalization, specificity, and inference; suggesting our method accurately captures the dynamics of shape over time.

%% file: methods.tex
\section{Methods}

\subsection{Notation}
Given a cohort of $N$ subjects, each subject has a time-sequence of $T-$shapes, where the temporal frames are consistent across subjects. Each shape is represented by a set of $d-$dimensional points (or particles).
In this work, shape is segmented from volumetric images, so $d=3$. 
We consider two forms of random variables: configuration and shape space variables, where the former captures sample-specific geometry and the latter describes population-level shape statistics.
The configuration space variable $\X_{n,t}$ represents the particle position on the $n-$th subject at the $t-$th time point, where $n \in [1,2, \dots N]$ and $t \in [1,2, \dots T]$.
$M-$realizations of this random variable defines the point set (or PDM) of the $n,t-$shape, $\x_{n,t} = \left[ \x_{n,t}^1, \x_{n,t}^2, \dots, \x_{n,t}^M \right] \in \mathbb{R}^{dM} \ (\x_{n,t}^m \in \mathbb{R}^{d}) $, where $\x_{n,t}^m$ is the vector of the $\{x,y,z\}$ coordinates of the $m-$th particle. 
Generalized Procrustes alignment is used to estimate a rigid transformation matrix $\T_{n,t}$ that can transform the particles in the local coordinate $\x_{n,t}^m$ in the configuration space to the world common coordinate $\z_{n,t}^m$ in the shape space such that 
$\z_{n,t}^m = \T_{n,t}\x_{n,t}^m$.
In the cross-sectional PSM formulation, shape distribution in the shape space is modeled by a single random variable $\Z \in \mathbb{R}^{dM}$ that is assumed to be Gaussian distributed and only captures inter-subject variations.
We define two different shape space random variables, one represents shapes across subjects at a specific time point $t$ (i.e., inter-subject variable) and is denoted $\Z_t$ and the other represents shape across time for a specific subject $n$ (i.e., intra-subject variable) and is denoted $\Z_n$. 

\subsection{Proposed Spatiotemporal Optimization Scheme}

We build on the PSM approach for optimizing population-specific PDMs. Vanilla PSM assumes cross-sectional data (denoted hereafter ``cross-sectional PSM") and optimizes particle positions for observations at a single point in time using an entropy-based scheme \cite{cates2007shape, cates2017shapeworks}.
Here intuitively $\X_n=\X_{n,t=1}$ and $\Z=\Z_{t=1}$.
The optimization objective to be minimized is then defined as: 
\begin{equation}
    \mathcal{Q}_{cross-sectional} = \alpha H(\Z) - \sum_{n=1}^N H(\X_n) 
\end{equation}
where $H$ is the differential entropy of the respective random variable and $\alpha$ is a relative weighting parameter that defines the contribution of the correspondence objective $H(\Z)$ to the particle optimization process. In this work $\alpha$ was experimentally tuned to be initialized as $\alpha=100$ then gradually decreased during optimization until $\alpha=0.1$ for both the proposed and comparison method. 
Minimizing this objective balances two terms.
The first encourages a compact distribution of samples in the shape space, ensuring maximal correspondence between particles across shapes (\ie lower model complexity).  
The second encourages maximal uniformly-distributed spread of points across individual shapes so that shape is faithfully represented (\ie better geometric accuracy).

We propose a novel spatiotemporal optimization equation that disentangles the shape space entropy for $\Z_t$ and $\Z_n$. It is defined as follows:
\begin{equation}
    \mathcal{Q} = \alpha\left(\sum_{t=1}^T H(\Z_t) + \sum_{n=1}^N H(\Z_n)\right) - \sum_{n=1}^N \sum_{t=1}^{T}H(\X_{n,t})
    \label{eq:opt_obj}
\end{equation}
The first term encourages intra-subject correspondence across time points, the second inter-subject correspondence across sequences, and the third retains geometric accuracy across subjects and time points. 
The cost function is optimized via gradient descent.
To find the correspondence point updates, we must take the derivative of $H(\Z_n)$ and $H(\Z_t)$ with respect to particle positions.
Similarly to the cross-sectional PSM formulation \cite{cates2007shape}, we model $p(\Z_t)$ and $p(\Z_n)$ parametrically as Gaussian distributions with covariance matrices $\boldSigma_t$ and $\boldSigma_n$, respectively. 
These covariance matrices are directly estimated from the data.
The entropy terms are then given by:
\begin{equation}
    H(\Z_*) \approx \frac{1}{2}\log{\boldSigma_*} = \frac{1}{2} \sum_{i=1}^{dM} \log {\lambda_{*,i}} 
    \label{eq:HZ}
\end{equation}
 where $*$ represents either $t$ or $n$ and $\lambda_{*,i}$ are the eignevalues of $\boldSigma_*$.  
 We estimate the covariances from the data (closely following \cite{cates2007shape}) and find: 
\begin{equation}
    \frac{-\partial{H(\Z_*)}}{\partial{\X}} \approx  \Y_*(\Y_*^\top\Y_* + \alpha\I)^{-1}
    \label{eq:partialHZ}
\end{equation}
where the respective $\Y_*$ matrices denote the matrix of points minus the sample mean for the ensemble and the regularization $\alpha$ accounts for the possibility of diminishing gradients (see \cite{cates2007shape} for more detail).
Combining equation \ref{eq:partialHZ} with the shape-based updates explained in \cite{cates2007shape} we get an update for each point.

\subsection{Image-Based Comparison Method}
As a baseline for comparison, we consider the approach presented in ``An Image-based Approach for 3D Left Atrium Functional Measurements" \cite{morris2020image}. In this work, anatomic segmentations from  high-resolution Magnetic Resonance Angiographies (MRA) were registered and propagated through pairwise deformable registrations to cover the cardiac cycle.  We applied this to spatiotemporal SSM by fitting a PDM to a single corresponding time-point across the cohort, cross-sectionally, then individually propagating particles across time points for each subject using the image-to-image deformable registration transforms.  
While this approach provides intra- and inter-subject correspondences, the correspondence distribution across time points involves only one subject at a time and is unable to be optimized for shape statistics across the cohort.

\subsection{LDS Evaluation Metrics}\label{section:metrics}

A time-series generative model is required to evaluate how well a spatiotemporal PDM captures the underlying time dependencies. Here, we use a Linear Dynamic System (LDS).
LDS is equivalent to a Hidden Markov Model (HMM) except it employs linear Gaussian dynamics and measurements and uses a continuous latent/state variable instead of discrete (Fig. \ref{fig:SSM_variant}).
The LDS state representation, $\ldsS$, can be fit to the PDM, along with model parameters, via the expectation maximization (EM) algorithm. We provide a more detailed explanation of the LDS model and EM algorithm in the supplementary materials, Appendix A. 
Generalization and specificity of the fit LDS model are used to determine if the temporal dynamics are accurately portrayed by the PDM.

\vspace{-.1in}
\input{figures/LDS-fig} 
\vspace{-.1in}

\textbf{Partial and Full-Sequence Reconstruction.} If the time-dependency is captured, an LDS model fit on a subset of the PDM's should generalize to held out examples (via multi-fold validation). The reconstruction error is calculated as the root mean square error (RMSE) between unseen observation sequences $\x_{n,t}$ and those reconstructed from the state/latent space by the LDS model $\hat{\x}_{n,t}$:
\begin{equation}
    \text{RMSE} = \sqrt{\frac{1}{NTdM}\sum_{n=1}^N\sum_{t=1}^T\sum_{m=1}^M\sum_{i=1}^d\left(\x_{n,t}^{m_i} - \hat{\x}_{n,t}^{m_i}\right)^2}
    \label{eq:RMSE}
\end{equation}
Here $\hat{\x}_{n,t}$ is found by estimating the state values (using the E-step (A.2)) then applying the LDS observation equation (Eq. A.1.2) reconstruct observations. When provided a partial sequence, LDS should also be able to reconstruct the missing points. The LDS state equation (Eq. A.1.1) estimates the state values for randomly masked missing time points. The partial observations are then reconstructed in the same manner and compared to the full unmasked sequence.

\textbf{Specificity.} The specificity of the fit LDS model is another indication of how well the shape model captures the time-dependency. The fit LDS model should be specific to the training set, meaning it only represents valid instances of shape sequences. To test this, we sample new observations from the LDS model, then compute the RMSE between these samples and the closest training sequences. 

%% file: figures/LDS-fig.tex
\begin{figure}[h]
    \centering \begin {tikzpicture}[-latex ,auto ,node distance =1.25cm and 2cm ,on grid ,
    semithick ,
    state1/.style ={ circle ,top color =white , bottom color = processblue!20 ,
    draw,processblue , text=black ,minimum size =0.5 cm}, 
    state2/.style ={ circle ,color =white ,
    draw,processblue , text=black ,minimum size=0.5 cm}
    ]
    \node[state1] (x1){$\x_{n,1}$};
    \node[state2] (z1)[above=of x1]{$\s_{n,1}$};
    \node[state2] (z2)[right=of z1]{$\s_{n,2}$};
    \node[state1] (x2)[below=of z2, right= of x1]{$\x_{n,2}$};
    \node[draw=none,fill=none] (dots)[right=of z2]{$\cdots$};
    \node[state2] (zt)[right=of dots]{$\s_{n,T}$};
    \node[state1] (xt)[below=of zt]{$\x_{n,T}$};
    \path (z1) edge node[left] {$\W_1$} (x1);
    \path (z1) edge node[above] {$\A_2$} (z2);
    \path (z2) edge node[left] {$\W_2$} (x2);
    \path (z2) edge node[above] {$\A_3$} (dots);
    \path (dots) edge node[above] {$\A_T$} (zt);
    \path (zt) edge node[left] {$\W_T$} (xt);
    \end{tikzpicture}
    \caption{Time-varying LDS model}
    \label{fig:SSM_variant}
\end{figure}
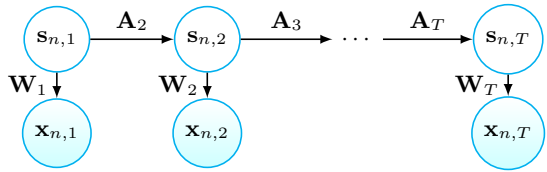

%% file: results.tex
\section{Results}
\subsection{4D Left Atrium Data}
\begin{figure}[ht!]
    \begin{center}
        \includegraphics[width=\textwidth]{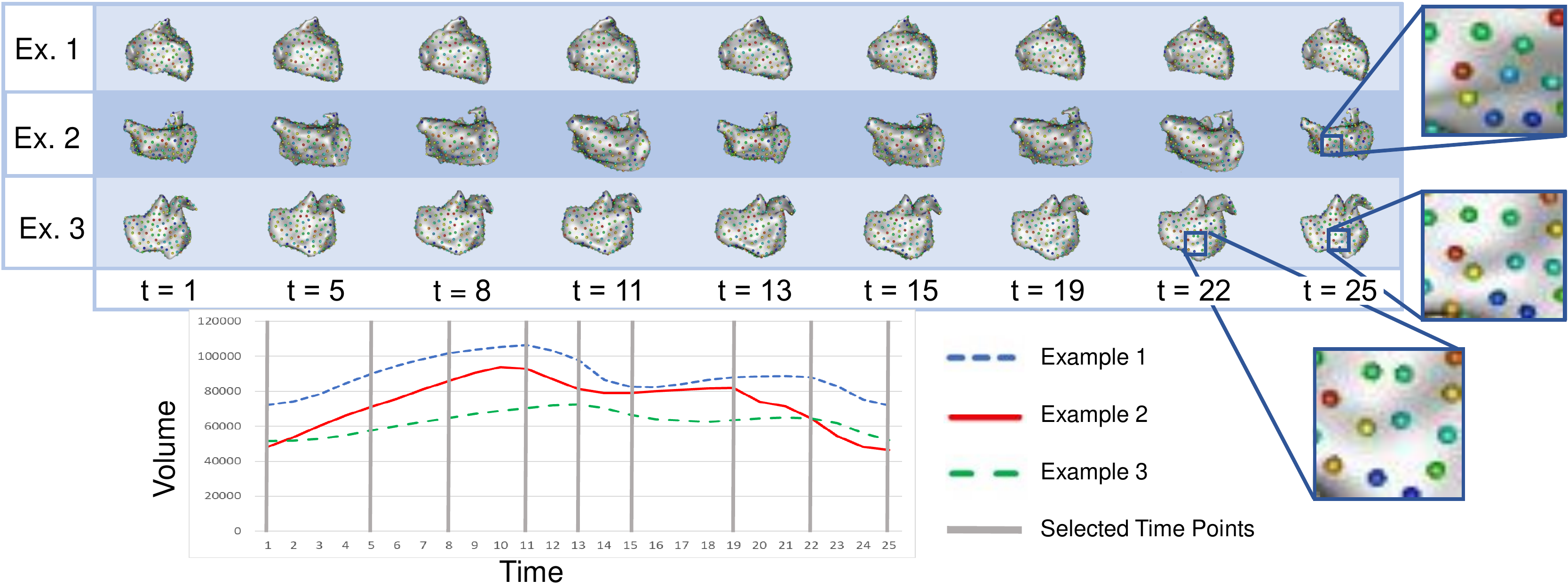}
        \caption{\textbf{PDM Examples} Correspondence points for three subjects on the ground truth meshes (anterior view). Color denotes correspondence (see zoomed in boxes). The plot of left atrium volume over time shows the time points selected for display.}
        \label{fig:examples}
    \end{center}
\end{figure}

We performed experimental analysis on a cohort comprised of 3D LGE and stacked CINE CMR scans through the left atrium for 28 patients presenting with atrial fibrillation between 2019 and 2020. Average patient age was 64.9 years with 15 male and 13 female. Scans were captured for each patient before and after a RF cardiac ablation procedure for a total of 56 scans. Each CINE scan contained 25 consistent time points covering the cardiac cycle. The temporal dimension was normalized at time of acquisition to cover one heart beat for each patient.  The number of milliseconds covered by the temporal dimension thus varies by patient. The 3D LGE images were manually segmented by a cardiac imaging expert and this segmentation was matched to the closest CINE time-point based on CMR trigger time. 
The segmentation was then transformed to each time point through time point to time point deformable registrations to create a full 3D segmentation sequence \cite{morris2020image}.
This dataset is selected to demonstrate the robustness of the proposed method as the left atrium shape varies greatly across patients and atrial fibrillation effects the dynamics in differing ways (Fig. \ref{fig:examples}).

\vspace{-.16in}

\subsection{Evaluation}
We leverage ShapeWorks\cite{ShapeWorks}, the open source implementation of the PSM method, as a starting point for out implementation.
ShapeWorks utilizes particle splitting, constraining the number of particles to be a power of two. 
We chose to use 256 particles for both the proposed and comparison PDM, as this is the smallest number that is dense enough to accurately represent the geometry of the left atrium. 
Fig. \ref{fig:examples} displays the part of the PDM generated via the proposed method on three subjects at selected time points, illustrating both intra- and inter-subject correspondence. 
The subset of time points selected for display were chosen to be meaningful points in the cardiac cycle using left atrium volume. 

\begin{figure}[ht!]
    \begin{center}
        \includegraphics[width=\textwidth]{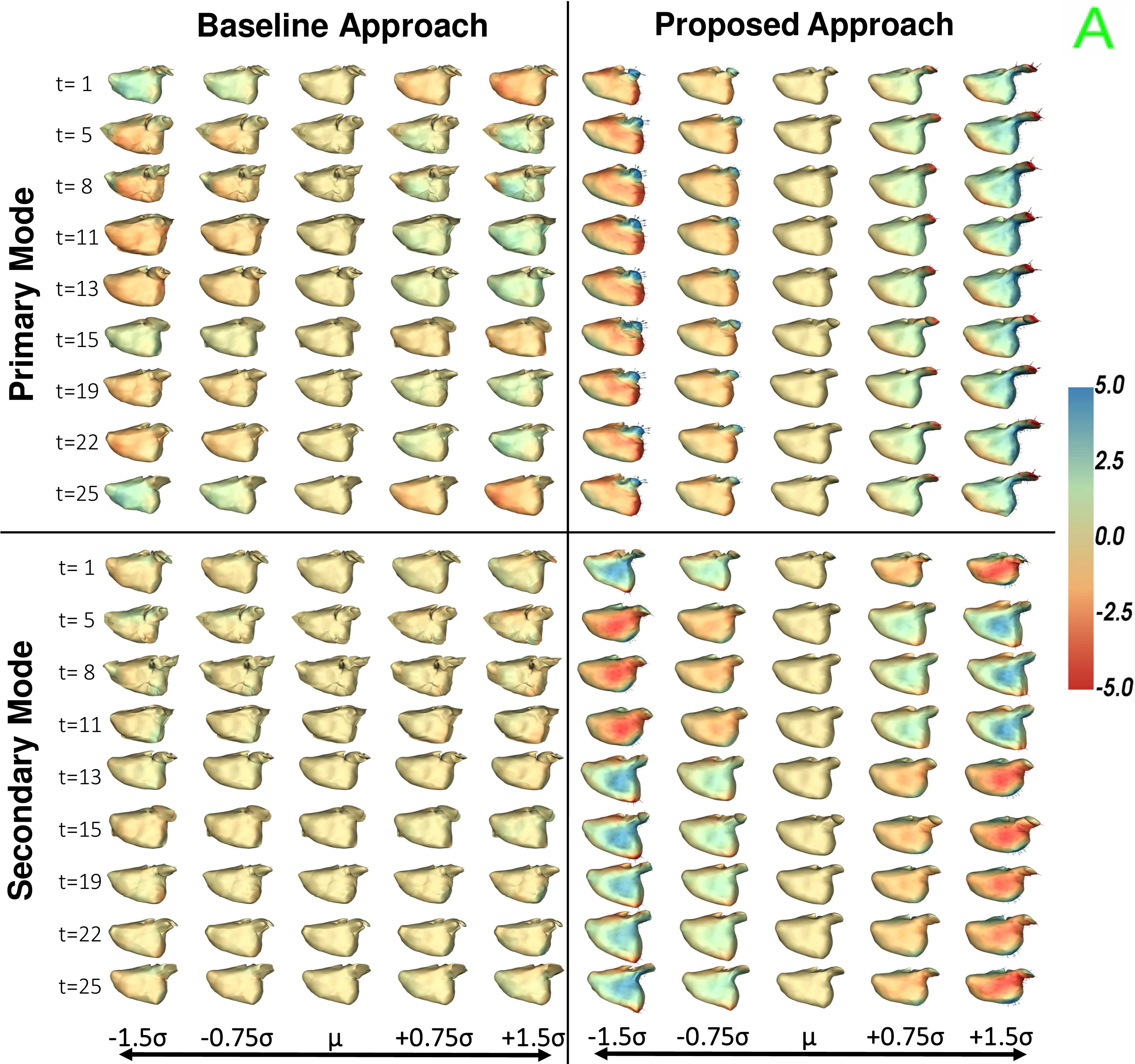}
        \caption{\textbf{Modes of Variation} Anterior view of Left Atrium across the primary and secondary mode of variation. Heat maps show the distance from mean shape.}
        \label{fig:modes}
    \end{center}
\end{figure}

Particle correspondence captures the modes of variation present in the population. Fig. \ref{fig:modes} displays the primary and secondary modes of variation captured by the two shape models. In the proposed shape model, the primary mode of variation captures the left atrium appendage elongation and the secondary captures the sphericity of the left atrium. This is to be expected as this is the main source of variation across patients and time. The shape variation captured in the modes of the baseline approach PDM has smaller magnitude and is more difficult to interpret, suggesting weaker correspondence. Additionally, we observe the temporal dynamics are more smoothly captured by the proposed approach shape model. 
This example illustrates the utility of our approach as a tool for statistically quantifying and visualizing population variation.

We employ the LDS model to analyze how well the PDM's capture the underlying dynamics. We fit LDS models with the same initial parameters latent dimension $L=64$ to both PDM's for 50 EM iterations. To ensure a fair comparison, training was done on a version of the correspondence particles ($d=3$) scaled uniformly to a range of 0 to 1. Thus, generalization and specificity evaluation metrics are interpreted on a relative scale.

\begin{figure}[ht!]
    \begin{center}
         \includegraphics[width=\textwidth]{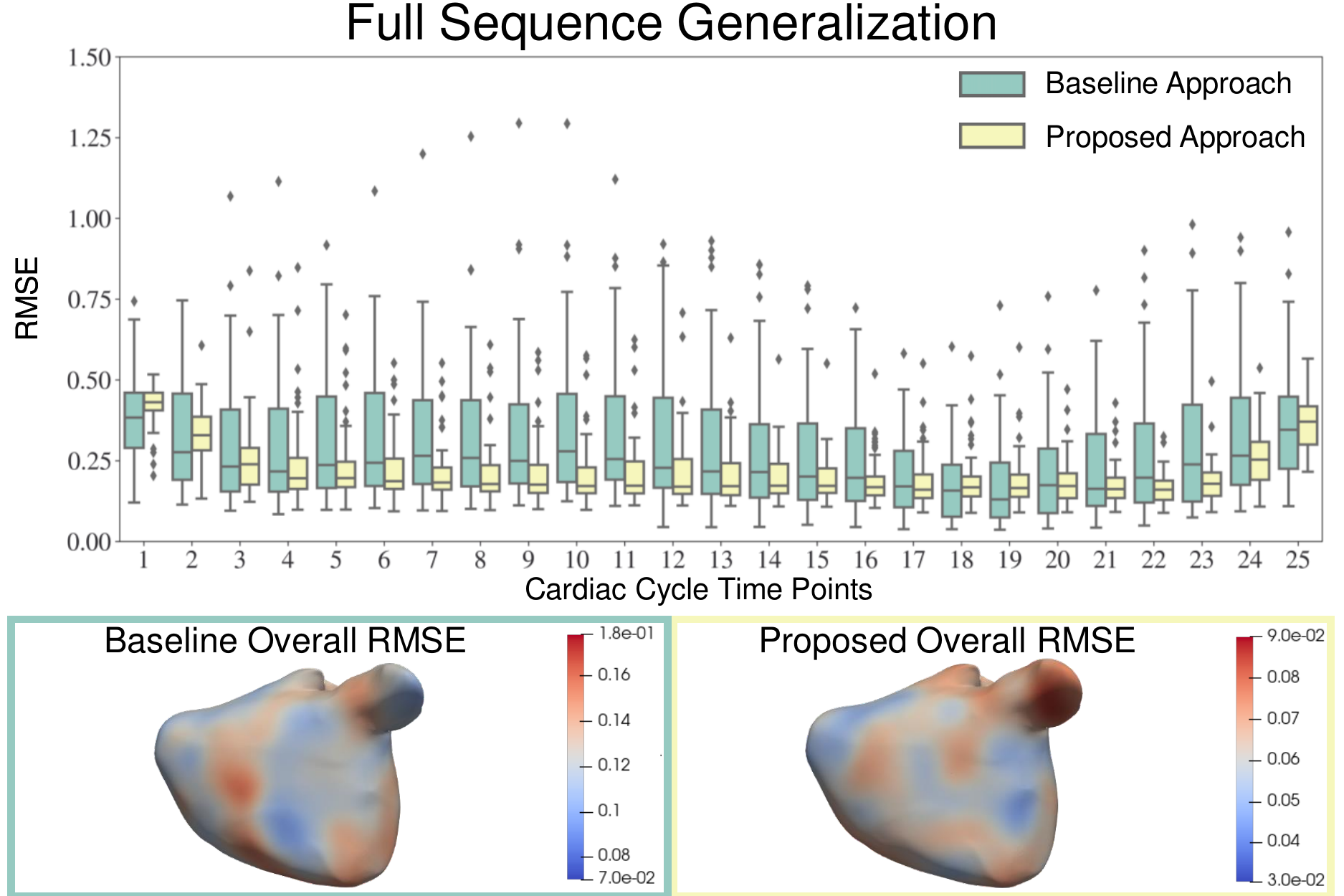}
        \caption{\textbf{Full Sequence Reconstruction} The box plot shows the distribution of particle-wise reconstruction RMSE at each time point. Below the average particle RMSE over all time points is displayed as a heat maps on a representative mesh, illustrating regional accuracy. Note the heat maps are scaled differently to make local changes visible.} \label{fig:full}  
    \end{center}
\end{figure}

\begin{figure}[ht!]
    \begin{center}
         \includegraphics[width=\textwidth]{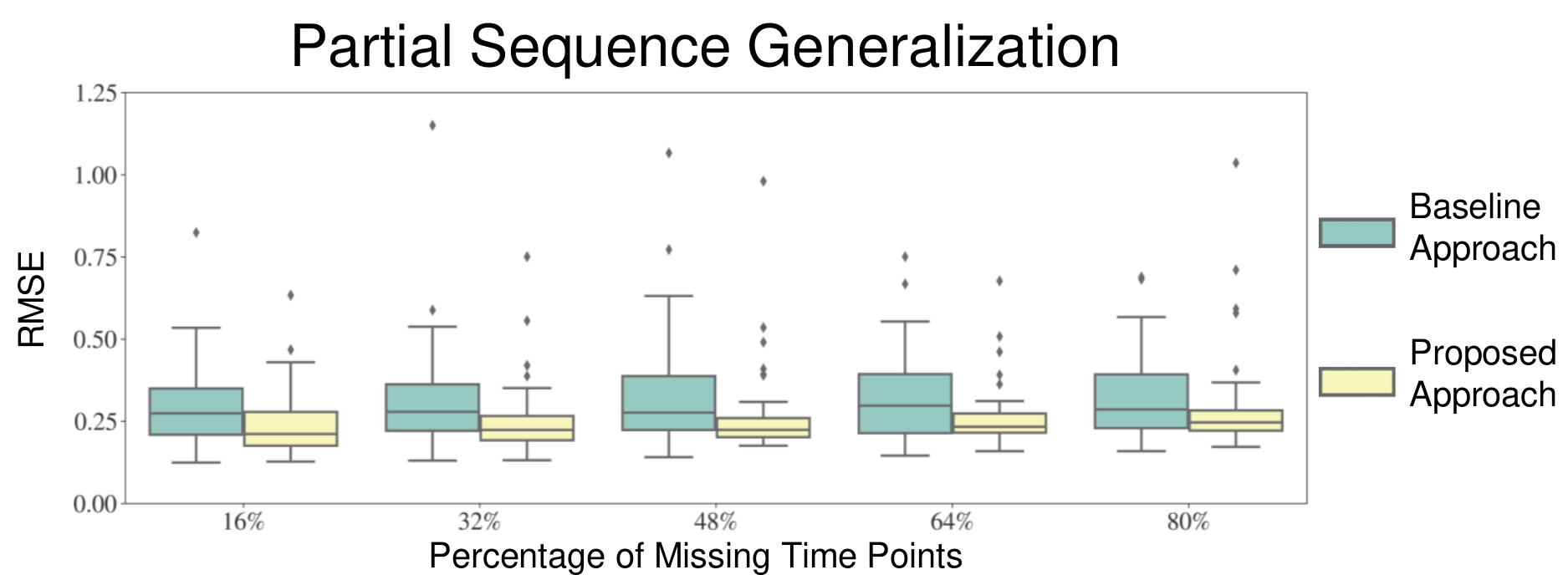}
        \caption{\textbf{Partial Sequence Reconstruction} The box plot shows the distribution of error with various percentages of missing time points.} \label{fig:partial}  
    \end{center}
\end{figure}

\begin{figure}[ht!]
    \begin{center}
         \includegraphics[width=\textwidth]{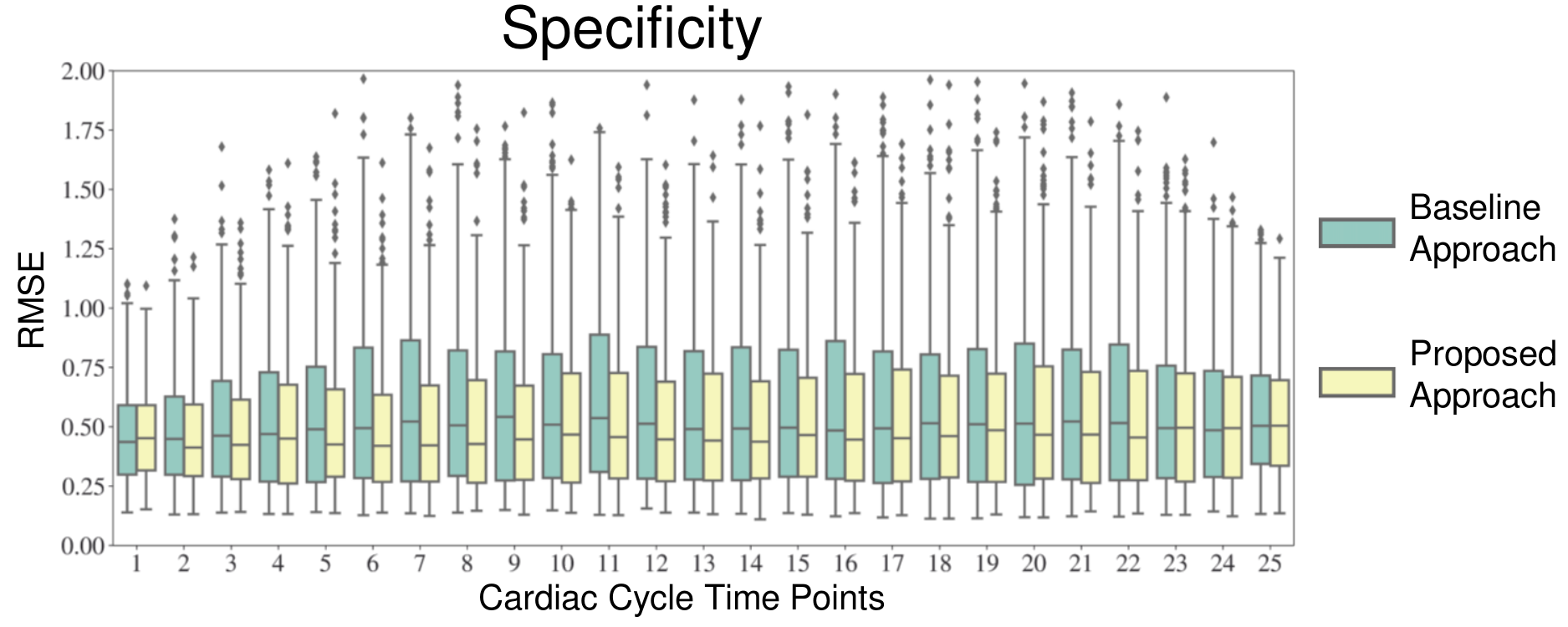}
        \caption{\textbf{Specificity} The box plot shows the distribution of particle-wise RMSE between sampled and closest true particle sets at each time point.} \label{fig:specificity}  
    \end{center}
\end{figure}

Fig. \ref{fig:full} displays the distribution of full sequence reconstruction error across time points. We can see that LDS fit to the proposed PDM more accurately reconstructs held out sequences.  
The partial sequence generalization results in Fig. \ref{fig:partial} similarly demonstrates that LDS fit to our PDM more accurately infers missing time points within a sequence and notably performs better with a larger percent missing. Finally, the specificity results, shown in Fig. \ref{fig:specificity} demonstrate that our approach leads to a more specific LDS model.

Table \ref{table:lds} provides the overall RMSE with standard deviation for these metrics generated by 5-fold cross-validation experiments on the LA data. Lower reconstruction error on held out sequences (full or partial) indicates the LDS model has better generalization, suggesting the underlying shape dynamics are better represented. Regarding specificity, a lower RMSE indicates that the generated sequences sampled from LDS are closer to the training sequences, suggesting the time-dependency is accurately described by the PDM. Thus these metrics support the conclusion that our PSM technique is capable of accurately capturing the temporal projection of shapes.

\begin{table}[ht!]
    \begin{center}
    \caption{Overall RMSE for LDS Metrics.}
    \begin{tabular}{|l|c|c|c|}
    \hline
    \textbf{Approach} & \textbf{Full-Seq. Generalization} & \textbf{Partial-Seq. Generalization} &  \textbf{Specificity} \\
    \hline
    Baseline & $0.345 \pm 0.065$ & $0.329 \pm 0.233$ & $0.576 \pm 0.357$ \\ \hline
    Proposed & $0.243 \pm 0.035$ & $0.270 \pm 0.204$ & $0.519 \pm 0.296$ \\ \hline
    \end{tabular}\label{table:lds}
    \end{center}
\end{table}

%% file: conclusion.tex
\section{Conclusion}
We introduced a novel PDM optimization scheme for spatiotemporal SSM.
Our method ensures intra-subject correspondence across time points, inter-subject correspondence across sequences, and geometric accuracy across both subjects and time points.
We demonstrated the efficacy of our method in capturing population variability in the left atrium throughout the cardiac cycle. 
Our method outperformed a comparison image-based approach for generating a spatiotemporal PDM in terms of capturing shape variation and the underlying time-dependency, as demonstrated via LDS generalization and specificity.
In future analysis, spatiotemporal SSM could be used to analyze group differences in the cardiac cycle of patients before and after the ablation procedure. 
Another consideration is that while our optimization objective disentangles subject and time correspondence, it does not fully capture their interaction as a time-sequence generative modeling approach would.
In future work, such a generative model could be integrated into the optimization scheme so that the modeled distribution informs particle position updates. 

\subsubsection{Acknowledgements}
This work was supported by the National Institutes of Health under grant numbers NIBIB-U24EB029011, NIAMS-R01AR076120, \\ NHLBI-R01HL135568, and  NIBIB-R01EB016701.
The content is solely the responsibility of the authors and does not necessarily represent the official views of the National Institutes of Health.
The authors would like to thank the University of Utah Division of Cardiovascular Medicine for providing left atrium MRI scans and segmentations from the Atrial Fibrillation projects.
%